\RequirePackage{etoolbox}
\csdef{input@path}%
{%
 {sty/},
 {img/},
}
\documentclass{elsevierbook}

\newcommand{\vect}[1]{\boldsymbol{#1}}
\usepackage{systeme}
\usepackage{gensymb}
\usepackage{amsmath,cases}
\mathchardef\mhyphen="2D 
\usepackage{float}
\usepackage{siunitx}
\usepackage{textcomp}
\usepackage[ruled]{algorithm2e}
\usepackage{multirow}
\usepackage{multicol}

\usepackage{pbox}
\usepackage{slashbox}
\usepackage{diagbox}
\usepackage{cancel}
\usepackage{array}
\usepackage{tabularx}
\usepackage{makecell}

\newcommand{\PreserveBackslash}[1]{\let\temp=\\#1\let\\=\temp}
\newcolumntype{C}[1]{>{\PreserveBackslash\centering}p{#1}}
\newcolumntype{R}[1]{>{\PreserveBackslash\raggedleft}p{#1}}
\newcolumntype{L}[1]{>{\PreserveBackslash\raggedright}p{#1}}

\usepackage{xcolor}

\newcommand{\review}[1]{\textcolor{black}{#1}}

\usepackage{natbib}
\begin{document}


  
\begin{frontmatter}

\chapter{GelTip Tactile Sensor for Dexterous Manipulation in Clutter}\label{chap1}
\begin{aug}
\author[addressrefs={ad1}]%
{%
\fnm{Daniel} \snm{Fernandes Gomes}%
}%
\author[addressrefs={ad1}]%
{%
\fnm{Shan} \snm{Luo}%
}%
\address[id=ad1]%
{%
smARTLab,
Department of Computer Science,
University of Liverpool,
United Kingdom. \\
Emails: \{danfergo, shan.luo\}@liverpool.ac.uk.
}%

\end{aug}

\begin{abstract}
Tactile sensing is an essential capability for robots that carry out dexterous manipulation tasks. While cameras, Lidars and other remote sensors can assess a scene globally and instantly, tactile sensors can reduce their measurement uncertainties and gain information about the local physical interactions between the in-contact objects and the robot, that are  often not accessible via remote sensing.
Tactile sensors can be grouped into two main categories: electronic tactile skins and camera based optical tactile sensors. The former are slim and can be fitted to different body parts, whereas the latter assume a more prismatic shape and have much higher sensing resolutions, offering a good advantage for being used as robotic fingers or fingertips.  
One of such optical tactile sensors is our \textit{GelTip} sensor that is shaped as a finger and can sense contacts on any location of its surface. As such, the \textit{GelTip} sensor is able to detect contacts from all the directions, like a human finger. To capture these contacts, it uses a camera installed at its base to track the deformations of the opaque elastomer that covers its hollow, rigid and transparent body. Thanks to this design, a gripper equipped with \textit{GelTip} sensors is capable of simultaneously monitoring contacts happening inside and outside its grasp closure. Experiments carried out using this sensor demonstrate how contacts can be localised, and more importantly, the advantages, and even possibly a necessity, of leveraging all-around touch sensing in dexterous manipulation tasks in clutter where contacts may happen at any location of the finger. In particular, experiments carried out in a Blocks World environment show that the detected contacts on the fingers can be used to adapt planned actions during the different moments of the reach-to-grasp motion. All the materials for the fabrication of the \textit{GelTip} sensor can be found at https://danfergo.github.io/geltip/\end{abstract}
\begin{keywords}
\kwd{Tactile sensors}
\kwd{Dexterous manipulation}
\kwd{GelTip sensor}
\end{keywords}

\end{frontmatter}

\section{Introduction}
\label{sec:sample1}

As humans, robots need to make use of tactile sensing when performing dexterous manipulation tasks in cluttered environments such as at home and in warehouses. In such cases, the positions and shapes of objects are uncertain, and it is of critical importance to sense and adapt to the cluttered scene. With cameras, Lidars and other remote sensors, large areas can be assessed instantly~\cite{peel2018localisation}. However, measurements obtained using such sensors often suffer from large uncertainties, occlusions and variance of factors like light conditions and shadows. Thanks to the direct interaction with the object, tactile sensing can reduce the measurement uncertainties of remote sensors and it is not affected by the changes of the aforementioned surrounding conditions. Furthermore, tactile sensing gains information of the physical interactions between the objects and the robot end-effector that is often not accessible via remote sensors, e.g., incipient slip, collisions and detailed geometry of the object. As dexterous manipulation requires precise information of the interactions with the object, especially in moments of in-contact or near-contact, it is of crucial importance to attain these accurate measurements provided by tactile sensing. For instance, failing to estimate the size of an object by \SI{1}{\mm}, or its surface friction coefficient, during (and also right before) a grasp might result in severely damaging the tactile sensor or dropping the object. In contrast, failing to estimate the object shape by a few centimeters farther away, will not make a big impact on the manipulation. To this end, camera vision and other remote sensors can be used to produce initial estimations of the object and plan manipulation actions, whereas tactile sensing can be used to refine such estimates and facilitate the in-hand manipulation~\cite{luo2017robotic,luo2021vitac}.

The usage of tactile sensors for manipulation tasks has been studied since \cite{opticalSensorsDexterousManipulation} and in the past years a wide range of tactile sensors and working principles have been studied in the literature~\cite{dahiya2013directions,luo2017robotic,luo2021vitac}, ranging from flexible electronic skins~\cite{kaltenbrunner2013ultra}, fiber optic based sensors~\cite{xie2013fiber}, capacitive tactile sensors~\cite{maiolino2013flexible}, to camera based optical tactile sensors~\cite{TacTipFamily,GelSight2017}, many of which have been employed to aid robotic grasping~\cite{kappassov2015tactile}. Electronic tactile skins and flexible capacitive tactile sensors can be adapted to different body parts of the robot that have various curvatures and geometry shapes. However, due to the necessary of dielectrics for each sensing element, they produce considerably low resolution tactile readings. For example, a WTS tactile sensor from Weiss Robotics used in~\cite{luo2015novel,luo2016iterative,luo2019iclap} has 14 $\times$ 6 taxels (tactile sensing elements). In contrast, camera based optical tactile sensors provide higher-resolution tactile images. However, on the other side, they usually have a bulkier shape due to the requirement of hosting the camera and the gap between the camera and the tactile membrane.

\begin{figure}[t]
\centering
\includegraphics[width=\linewidth]{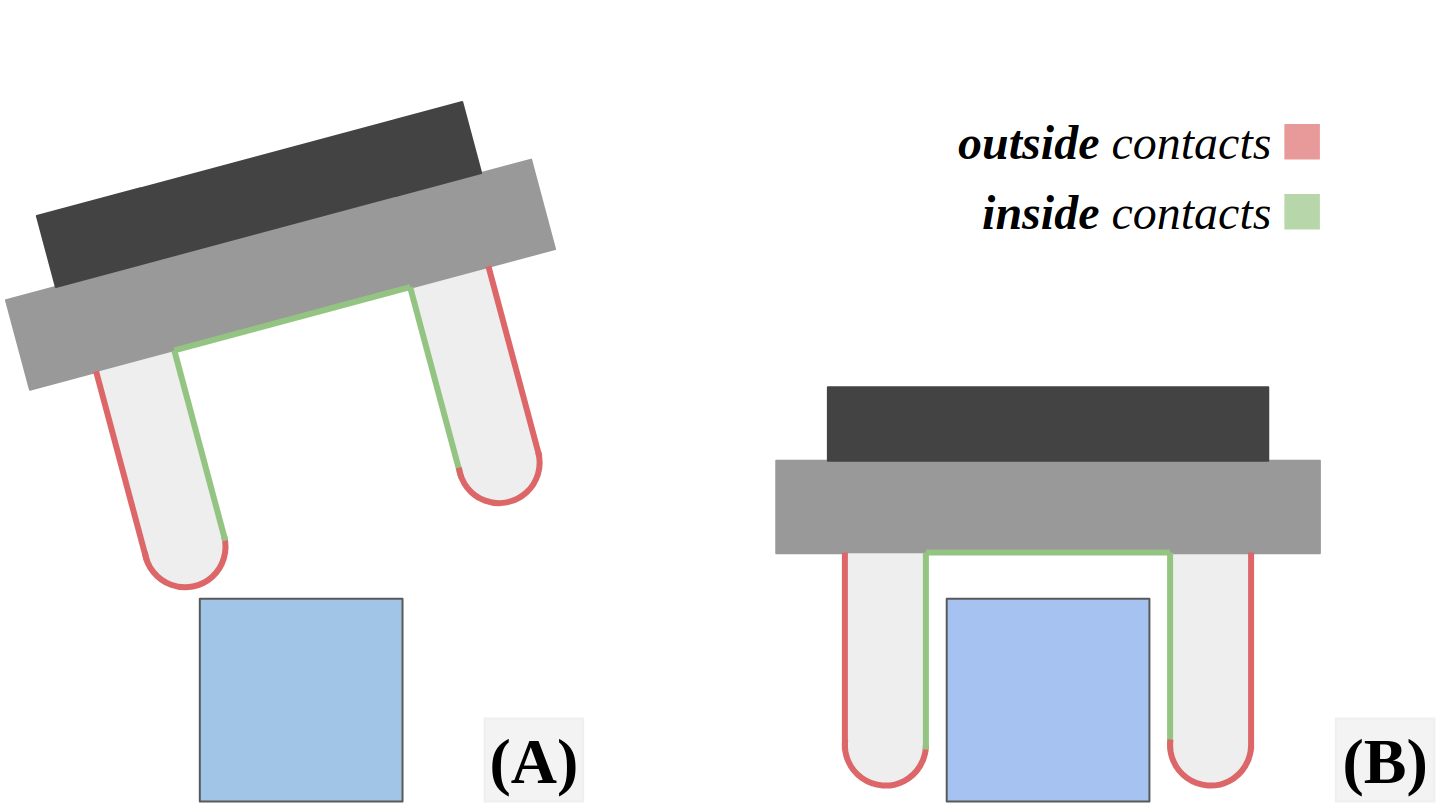} \\
\caption{There are two distinct areas of contact highlighted in the robot gripper during a manipulation task: \textbf{(A)} Outside contacts when the robot is probing or steering the object to be grasped; \textbf{(B)} Inside contacts when the object is within the grasp closure, which can guide the grasping. }
\label{fig:contact_areas}
\end{figure}

Optical tactile sensors can be grouped in two main groups: marker-based and image-based, with the former being pioneered by the \textit{TacTip} sensors~\cite{TacTip2009} and the latter by the \textit{GelSight} sensors ~\cite{RetrographicSensing}. As the name suggests, marker-based sensors exploit the tracking of markers printed on a soft domed membrane to perceive the membrane displacement and the resulted contact forces. By contrast, image-based sensors directly perceive the raw membrane with a variety of image recognition methods to recognise textures, localise contacts and reconstruct the membrane deformations, etc. Because of the different working mechanisms, marker-based sensors measure the surface on a lower resolution grid of points, whereas image-based sensors make use of the full resolution provided by the camera. Some \textit{GelSight} sensors have also been produced with markers printed on the sensing membrane \cite{GelSight2017GeometrySlip}, enabling marker-based and image-based methods to be used with the same sensor. Both families of sensors have been produced with either flat sensing surfaces or domed/finger-shaped surfaces.

In this chapter, we will first review existing optical tactile sensors in Section~\ref{sec:overview}, and then we'll look in detail into one example of such image-based tactile sensors, i.e., the \textit{GelTip}~\cite{geltip,gomes2020geltip}, in Section~\ref{sec:sensormodel}. The \textit{GelTip} is shaped as a finger, and thus it can be installed on traditional and off-the-shelf grippers to replace its fingers, and enable contacts to be sensed inside and outside the grasp closure that are shown in Figure~\ref{fig:contact_areas}. In Section~\ref{sec:experimentresults}, we will look into experiments carried out using the \textit{GelTip} sensor that demonstrate how contacts can be localised, and more importantly, the advantages, and possibly a necessity, of leveraging all-around touch sensing in dexterous manipulation tasks in clutter. In particular, experiments carried out in a Blocks World environment show that the detected contacts on the fingers can be used to adapt planned actions during the different moments of the reach-to-grasp motion.

 
\section{An overview of the tactile sensors}
\label{sec:overview}

\review{
Compared to remote sensors like cameras, tactile sensors are designed to assess the properties of the objects via physical interactions, e.g., geometry, texture, humidity and temperature. A large range of working principles have been actively proposed in the literature in the past decades~\cite{dahiya2013directions,luo2017robotic,dahiya2009tactile}. An optical tactile sensor uses a camera enclosed within its shell and pointing at its tactile membrane (an opaque window membrane made of a soft material), to capture the properties of the objects, from the deformations caused to its tactile membrane by the in contact object. Such characteristics ensure that the captured tactile images are not affected by the external illumination variances. To perceive the elastomer deformations from the captured tactile images, multiple working principles have been proposed. We group such approaches in two categories: marker tracking and raw image analysis. Optical tactile sensors contrast to electronic tactile skins that usually have lower thickness and are less bulky. They are flexible and can adapt to different body parts of the robot that have various curvatures and geometry shapes. However, each sensing element of most of the tactile skins, e.g., a capacitive transducer, has the size of a few square millimetres or even centimetres, which results in a limited spatial resolution of the tactile skins. Here we do not cover such skins as these are an extensive topic on its own, however, we point the reader to two surveys that extensively cover these sensors \cite{DexterousTactileSensorsSurvey, electronicSkins}.
}

 


\subsection{Marker-based optical tactile sensors}
The first marker-based sensor proposal can be found in \cite{gelforce}, however, more recently an important family of marker-based tactile sensors is the TacTip family of sensors described in \cite{TacTipFamily}. Since its initial domed shaped version \cite{TacTip2009}, different morphologies have been proposed, including the TacTip-GR2 \cite{tactipGR2} of a smaller fingertip design, TacTip-M2 \cite{tactipM2} that mimicks a large thumb for in-hand linear manipulation experiments, and TacCylinder to be used in capsule endoscopy applications. Thanks to their miniaturised and adapted design, TacTip-M2~\cite{tactipM2} and TacTip-GR2~\cite{tactipGR2} have been used as fingers (or finger tips) in robotic grippers. Although each TacTip sensor introduces some manufacturing improvements or novel surface geometries, the same working principle is shared: white pins are imprinted onto a black membrane that can then be tracked using computer vision methods. 

As shown in Table~\ref{table:rl_summary_1}, there are also other optical tactile sensors that track the movements of markers. In~\cite{FingerVision}, an optical tactile sensor named FingerVision is proposed to make use of a transparent membrane, with the advantage of gaining proximity sensing. However, the usage of the transparent membrane makes the sensor lack the robustness to external illumination variance associated with touch sensing. In~\cite{ColorMixingTactileSensor}, semi-opaque grids of magenta and yellow makers, painted on the top and bottom surfaces of a transparent membrane are proposed, in which the mixture of the two colours is used to detect horizontal displacements of the elastomer. In \cite{greendots}, green florescent particles are randomly distributed within the soft elastomer with black opaque coating so that a higher number of markers can be tracked and used to predict the interaction with the object, according to the authors. In \cite{greenDotsMulti}, a sensor with the same membrane construction method, 4 Raspberry PI cameras and fisheye lenses has been proposed for optical tactile skins.

\begin{table}
\centering
\caption{A summary of influential Marker-based optical tactile sensors}
\def\arraystretch{1.5}
\begin{tabular}{R{0.15\linewidth} | p{0.23\linewidth} p{0.46\linewidth} }   & \textbf{Sensor Structure} & \textbf{Illumination and Tactile Membrane} \\ 

\hline
 
\textbf{TacTip} 
\cite{TacTip2009} & The \textit{TacTip} has a domed (finger) shape, $40 \times 40 \times 85$ mm, and tracks 127 pins. It uses the Microsoft LifeCam HD webcam. & \multirow{2}{*}{\parbox{\linewidth}{The membrane is black on the outside, with white pins and filled with transparent elastomer inside. Initially the membrane was cast from VytaFlex 60 silicone rubber, the pins painted by hand and the tip filled with the optically clear silicone gel (Techsil, RTV27905); however, currently the entire sensor can be 3d-printed using a multi-material printer (Stratasys Objet 260 Connex), with the rigid parts printed in Vero White
material and the compliant skin in the rubber-like TangoBlack+. }} \\
\textbf{TacTip-M2} \cite{tactipM2} & It has a thumb-like or semi-cylindrical shape, with TacTip-M2 $ 32 \times 102 \times 95$~mm and it tracks  80 pins.  \\

\textbf{TacTip-GR2} \cite{tactipGR2} & It has a cone shape with a flat sensing membrane, and is smaller than the TacTip, $40 \times 40 \times 44$ mm, tracks 127 pins and uses the Adafruit SPY PI camera.    \\ 

\textbf{TacCylinder} 
\cite{tacCylinder}
& A catadioptric mirror is used to track the 180 markers around the sensor cylindrical body.   \\ \hline

\textbf{FingerVision} \cite{FingerVision} & 
It uses a ELP Co. USBFHD01M-L180 camera with an 180 degree fisheye lens. It has approximately  $40 \times 47 \times 30 $ mm. & The membrane is transparent, made with Silicones Inc. XP-565, with \SI{4}{\mm} of thickness and markers spaced by \SI{5}{\mm}. No internal illumination is used, as it the membrane transparent. \\ \hline

\textbf{Subtractive Color Mixing}
\cite{ColorMixingTactileSensor} &N/A & Two layers of  semi-opaque colored markers is proposed. SortaClear 12 from Smooth-On, clear and with Ignite pigment, is used to make the inner and outer sides. \\ \hline

\textbf{Green Markers}
\cite{greendots} & 
The sensor has a flat sensing surface, measures $50 \times 50 \times 37 $~mm and is equipped with a ELP USBFHD06H RGB camera with a fisheye lens.
& \multirow{2}{*}{\parbox{\linewidth}{It is composed of three layers: stiff elastomer, soft elastomer with randomly distributed green florescent particles in it and black opaque coating. The stiff layer is made of ELASTOSIL® RT
601 RTV-2 and is poured directly on top of the electronics, the soft layer is made of Ecoflex™ GEL (shore hardness 000-35) with the markers mixed in, and the final coat layer is made of ELASTOSIL® RT 601 RTV-2 (shore hardness 10A) black silicone. A custom board with an array of SMD white LEDs is mounted on the sensor base, around the camera. }} \\ 

\textbf{Multi-camera Skin} \cite{greenDotsMulti} & 
It has a flat prismatic shape of $49\times51\times17.45$ mm.
Four Pi cameras are assembled in a $2\times2$ array and fish-eye lenses are used to enable its thin shape. &

\end{tabular}
\label{table:rl_summary_1}
\end{table}

\subsection{Image-based optical tactile sensors}
On the other side of the spectrum, the GelSight sensors, initially proposed in~\cite{RetrographicSensing}, exploit the entire resolution of the tactile images captured by the sensor camera, instead of just tracking makers. Due to the soft opaque tactile membrane, the captured images are robust to external light variations, and capture information of the touched surface's geometry structure, unlike most conventional tactile sensors that measure the touching force.  Leveraging the high resolution of the captured tactile images, high accuracy geometry reconstructions are produced in \cite{GelSightSmallParts, luo2018vitac,lee2019touching,cao2020spatio,lu2019surface,jiang2021vision}. In~\cite{GelSightSmallParts}, this sensor was used as fingers of a robotic gripper to insert a USB cable into the correspondent port effectively. However, the sensor only measures a small flat area oriented towards the grasp closure. In~\cite{gomes2019gelsight,gomes2021generation}, simulation models of the GelSight sensors are also created.

Markers were also added to the membrane of the GelSight sensors, enabling applying the same set of methods that were explored in the TacTip sensors. There are some other sensor designs and adaptations for robotic fingers in~\cite{GelSight2017, GelSlim, digit}. In \cite{GelSight2017}, matte aluminium powder was used for improved surface reconstruction, together with the LEDs being placed next to the elastomer, and the elastomer being slightly curved on the top/external side. In \cite{GelSlim}, the GelSlim is proposed, a design wherein a mirror is placed at a shallow and oblique angle for a slimmer design. The camera was placed on the side of the tactile membrane, such that it captures the tactile image reflected onto the mirror. A stretchy textured fabric was also placed on top of the tactile membrane to prevent damages to the elastomer and to improve tactile signal strength. Recently, an even more slim design has been proposed \SI{2}{\mm} \cite{gelslim3}, wherein an hexagonal prismatic shaping lens is used to ensure radially simetrically illumination. In \cite{digit}, DIGIT is also proposed with a USB ``plug-and-play'' port and an easily replaceable elastomer secured with a single screw mount.

In these previous works on camera based optical tactile sensors, multiple designs and two distinct working principles have been exploited. However, none of these sensors has the capability of sensing the entire surface of a robotic finger, i.e., both the sides and the tip of the finger. As a result, they are highly constrained in object manipulation tasks, due to the fact that the contacts can only be sensed when the manipulated object is within the grasp closure ~\cite{GelSightSmallParts, IncipientSlip, RegraspVisionTouch}. To address this gap, we propose the finger-shaped sensor named GelTip that captures tactile images by a camera placed in the center of a finger-shaped tactile membrane. It has a large sensing area of approximately \SI{75}{\cm\squared} (\textit{vs.} \SI{4}{\cm\squared} of the GelSight sensor) and a high resolution of 2.1 megapixels over both the sides and the tip of the finger, with a small diameter of \SI{3}{\cm} (\textit{vs.} \SI{4}{\cm} of the TacTip sensor). More details of the main differences between the GelSight sensors, TacTip sensors and our GelTip sensor are given in Table~\ref{table:rl_summary}.

\begin{table}
\centering
\caption{A summary of influential flat and finger-shaped GelSight sensors}
\def\arraystretch{1.5}
\begin{tabular}{R{1.5cm} | p{3cm} p{3cm} p{3cm} }   & \textbf{Sensor Structure} & \textbf{Illumination} & \textbf{Tactile Membrane} \\ 

\hline
 
\textbf{GelSight}\cite{GelSightSmallParts} & It has a cubic design with a flat square surface. A Logitech C310 (1280 $\times$ 720) camera is placed at its base pointing at the top membrane. &  Four LEDs (RGB and white) are placed at the base. The emitted light is guided by the transparent hard surfaces on the sides, so that it enters the membrane tangentially. &  A soft elastomer layer is placed on top of a rigid, flat and transparent acrylic sheet. It is painted using semi-specular aluminum flake powder.\\ 

\textbf{GelSight} \cite{GelSight2017} &  It has a close-to hexagonal prism shape. The used webcam is also the Logitech C310. & Three sets of RGB LEDs are positioned (close to) tangent to the elastomer, with a \SI{120}{\degree} angle from each other. & A matte aluminium powder is proposed for improved surface reconstruction. Its elastomer has a flat bottom and a curved top.  \\ 

\textbf{GelSlim} \cite{GelSlim} &  A mirror placed at a shallow oblique angle and a Raspberry Pi Spy (640 $\times$ 480) camera is used to capture the tactile image reflected by the mirror. & A single set of white LEDs is used. These are pointed at the mirror, so that the light is reflected directly onto the tactile membrane. & A stretchy and textured fabric on the tactile membrane prevents damages to the elastomer and results in improved tactile signal strength.\\

\textbf{GelSlim v3}\cite{gelslim3} & It is shaped similar to \cite{GelSightSmallParts, GelSight2017} however slimmer \SI{20}{\mm} of thickness, and a round sensing surface. & A custom hexagonal prism  is constructed to ensure radially symmetric illumination. & An elastomer with Lambertian reflectance is used, as proposed in \cite{GelSight2017}. \\ 

\textbf{DIGIT} \cite{digit} & A prismatic design, with curved sides. An OmniVision OVM7692 (640 $\times$ 480) camera is embedded in the custom circuit board. & Three RGB LEDs are soldered directly into the circuit board, illuminating directly the tactile membrane.  & The elastomer can be quickly replaced using a single screw mount. \\

\textbf{Round Fingertip} \cite{br2020soft}  & It has a round membrane, close to a quarter of sphere. A single  \SI{160}{\degree} FoV Raspberry Pi (640 $\times$ 480) is installed on its base. & Two rings of LEDs are placed on the base of the sensor, with the light being guided through the elastomer. & Both rigid and soft parts of the membrane are cast, using SLA 3D printed molds.  \\ 

\textbf{OmniTact} \cite{padmanabha2020omnitact} &  It has a domed shape. Five endoscope cameras (400 $\times$ 400) are installed on a core mount, and placed orthogonally to each other: pointing at the tip and sides. & RGB LEDs are soldered both onto the top and sides of the sensor. & The elastomer gel is directly poured onto the core mount (and cameras) without any rigid surface or empty space in between. \\

\textbf{GelTip} \cite{geltip} & It has a domed (finger) shape, similar to a human finger. A Microsoft Lifecam Studio webcam (1920 $\times$ 1080) is used.   & Three sets of LEDs, with a \SI{120}{\degree} angle from each other, are placed at the sensor base, and the light is guided through the elastomer.   &  An acrylic test tube is used as the rigid part of the membrane. The deformable elastomer is cast using a three-part SLA/FFF 3D printed mold.     

\end{tabular}
\label{table:rl_summary}
\end{table}

With their compact design, the GelTip \cite{geltip} and other GelSight \cite{GelSightSmallParts, GelSlim, gelslim3, digit, cao2021touchroller} sensors are candidate sensors to be mounted on robotic grippers. Recently, custom grippers built using the GelSight working principle have also been proposed \cite{wilson2020design, she2019cable}. Two recent works \cite{br2020soft, padmanabha2020omnitact} also address the issue of the flat surface of previous GelSight sensors. However, their designs have large differences to ours. In \cite{br2020soft}, the proposed design has a tactile membrane with a surface geometry close to a quarter of a sphere. As a consequence, a great portion of contacts happening on the regions outside the grasp closure is undetectable. In \cite{padmanabha2020omnitact}, this issue is mitigated by the use of five endoscope micro cameras looking at different regions of the finger. However, this results in a significant increase of cost for the sensor, according to the authors, approximately US\$3200 (\textit{vs.} only around US\$100 for ours).


\section{The GelTip sensor}
\label{sec:sensormodel}
\subsection{Overview}

\begin{figure}
\centering
\includegraphics[width=\linewidth]{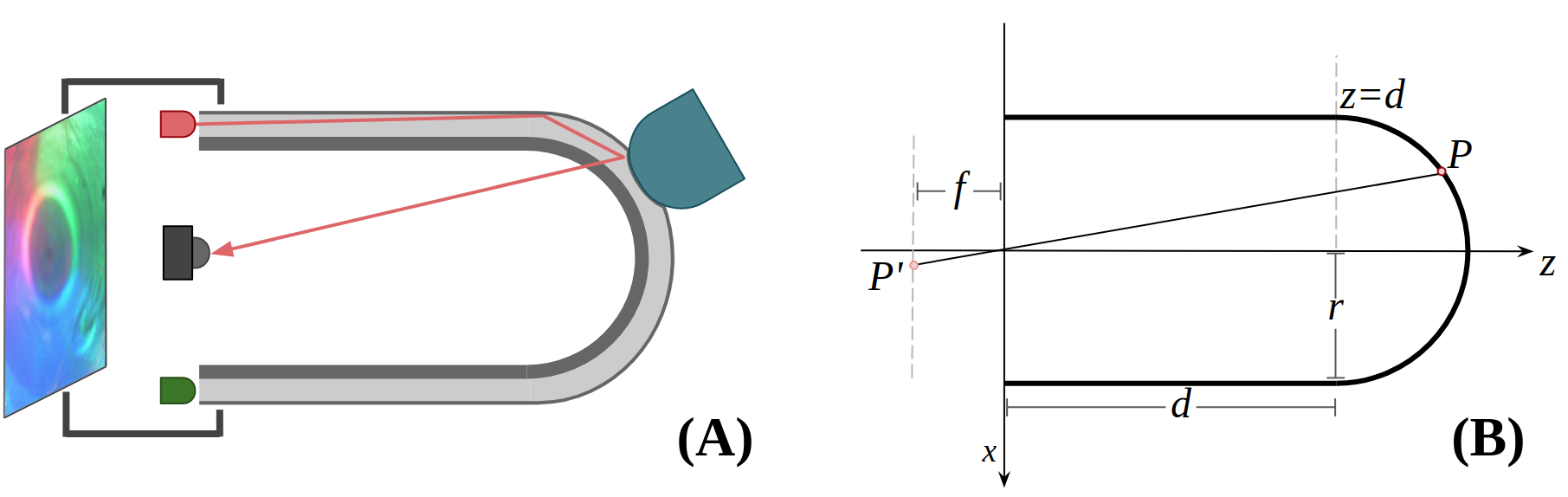} 
\caption{\textbf{(A)} The working principle of the proposed \textit{GelTip} sensor. The three-layer tactile membrane (rigid body, elastomer and paint coating) is shown in gray. The light rays emitted by the LEDs travel through the elastomer. As one object, shown in green, presses the soft elastomer against the rigid body, an imprint is generated. The resulted tactile image is captured by the camera sensor, placed in the core of the tactile sensor. An opaque shell, enclosing all the optical components, ensures the constant internal lighting of the elastomer surface. 
\textbf{(B)} Two-dimensional representation of the geometrical model of the \textit{GelTip} sensor. The tactile membrane is modeled as a cylindrical surface and a semi-sphere. An optical sensor of a focal-length $f$ is placed at the referential origin of the sensor, which projects a point on the surface of the sensor $P$ into a point $P'$ in the image plane. The sensor has a radius $r$ and its cylindrical body has a length $d$.}
\label{fig:sensor_innerworkings_and_model}
\end{figure}

As illustrated in Figure~\ref{fig:sensor_innerworkings_and_model}~(A), the GelTip optical tactile sensor is shaped as a finger, and its body consists of three layers, from the inside to the outer surface: a rigid transparent body, a soft transparent membrane and a layer of opaque elastic paint. In its base, a camera is installed, looking at inner surface of the cylinder. When an object is pressed against the tactile membrane, the elastomer distorts and indents according to the object shape. The camera can then capture the obtained imprint into a digital image for further processing. As the membrane is coated with opaque paint, the captured tactile images are immune to external illumination variances, which is characteristic of tactile sensing. To ensure that the imprint is perceptible from the camera view, LED light sources are placed adjacent to the base of the sensor, so that light rays are guided through the membrane. 

\subsection{The sensor projective model}
\label{subsec:sensor_projective_model}

For flat sensors, the relationship between the surface and the captured image can be often be easily obtained, or simply substituted by a scaling factor from single pixels to meters \cite{GelSightSmallParts, GelSight2017, IncipientSlip}. However, when considering highly curved sensors, it is important to study a more general projective function. In this subsection, we will look into how to derive such projective function $m$. As for the case of \textit{GelTip} sensor, $m$ maps pixels in the image space $(x',y')$ into points $ (x,y,z)$ on the surface of the sensor. Obtaining the protective function for other curved \textit{GelSight} sensors should be similar, requiring only sensor-specific adaptations. The camera is assumed to be placed at the referential origin, looking in the direction of $z$ axis. The sensor space takes the center of its base, which is also the center point of the camera, as the coordination origin $(0,0,0)$; the image space takes the center of the image as the origin $(0,0)$. Such a projection model is necessary for, among other applications, detecting the position of contacts in the 3D sensor surface. 

As illustrated in Figure~\ref{fig:sensor_innerworkings_and_model}~(B), the sensor surface can be modeled as a joint semi-sphere and an opened cylinder, both sharing the same radius $r$. The cylinder surface center axis and the $z$-axis are collinear, therefore, the center point of the semi-sphere can be set to $(0,0,d)$, where $d$ is the distance from the center point of the base of the semi-sphere to the center point of the base of the sensor. The location of any point on the sensor surface $(x,y,z)$ can be represented as follows: 

\begin{numcases}{}
  x^2 + y^2 + (z - d)^2 = r^2 & \text{for $z > d$} \\
  x^2 + y^2 = r^2 & \text{for $z <= d$}
\end{numcases}

By making the usual thin lens assumptions, the optical sensor is modeled as an ideal pinhole camera. The projective transformation that maps a point in the world space $ P $ into a point in the tactile image $P'$ can be defined using the general camera model~\cite{szeliski2011computer} as:

\begin{align}
\label{eq:camera_model}
P' &= K [\vect{R} | \vect{t}] P \\
K &= \begin{bmatrix} 
	\begin{array}{llll}
        fk & 0  & c_x & 0 \\
        0  & fl & c_y & 0 \\
        0  & 0  & 1   & 0 \\
    \end{array}    
\end{bmatrix}
\end{align}
where $P'=[x'z , y'z, z]^T$ is an image pixel and $P=[x,y,z,1]^T$ is a point in space, both represented in homogeneous coordinates here, $[R|t]$ is the camera's extrinsic matrix that encodes the rotation $ R $ and translation $ t $ of the camera, $K$ is the camera intrinsic matrix ($f$ is the focal length; $k$ and $l$ are the pixel-to-meters ratios; $c_x$ and $c_y$ are the offsets in the image frame). Assuming that the used camera produces square pixels, i.e., $k = l$, $fk$ and $fl$ can be replaced by $\alpha$, for mathematical convenience. 

The orthogonal projections in the $XZ$ and $YZ$ of a generic projection ray can be obtained by expanding the matrix multiplication given by Equation~\ref{eq:camera_model} and solving it w.r.t. $x$ and $y$:

 \begin{align}
\systeme*{
x'z = \alpha x + c_xz, 
y'z = \alpha y + c_yz,
z = z
} \Leftrightarrow
\systeme*{
\alpha x = x'z - c_xz,
\alpha y = y'z - c_yz
} \Leftrightarrow
\systeme*{
x = (\frac{x' - c_x}{\alpha})z,
y = (\frac{y' - c_y}{\alpha})z
}
\label{eq:proj_ray}
\end{align}

The desired mapping function $m: (x',y') \rightarrow (x,y,z) $ can then be obtained by constraining the $z$ coordinate through the intersection of the generic projection ray with the sensor surface, described in Equation~\ref{eq:function_f}, where $\chi=x'-c_x$ and $\gamma=y'-c_y$ and $\omega=\chi^2+\gamma^2$. The discontinuity region, i.e., a circumference, is found by setting $z=d$ in Equation \ref{eq:proj_ray}:

\begin{equation}
m(x',y') =
\left\{
\begin{array}{ll}
x &= \frac{\chi}{\alpha}z \\
y &= \frac{\gamma}{\alpha}z	\\	
z &= \left\{
\begin{array}{lll}
\sqrt{\frac{(r\alpha)^2}{\omega}} & \mbox{if } \omega  < (\frac{r\alpha}{d})^2 \\
\quad \\
\frac{{\alpha^2}2d + \sqrt{{(-{\alpha^2}2d)}^2 - 4\omega{{(d^2 - r^2)}{\alpha^2}}}}{2{(\omega + \alpha^2)}} & otherwise
\end{array}
\right. \\
\quad
\end{array}
\right.
\label{eq:function_f}
\end{equation}

\begin{figure}
  \centering
  \includegraphics[width=.5\linewidth]{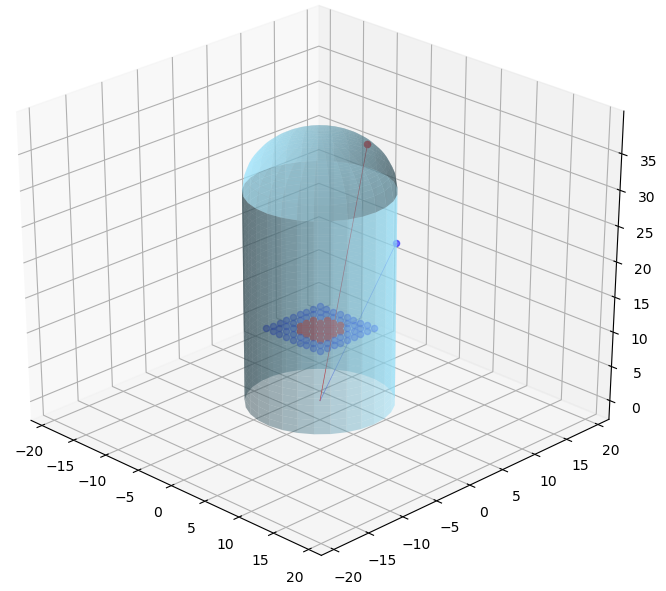} \\
   \caption{Two projection rays that correspond to the spherical (in red) and cylindrical (in navy blue) regions are depicted in the figure. Each ray intersects three relevant points: the frame of reference origin, a point in the sensor surface and the corresponding projected point in the image plane.}
  \label{fig:sim_model}
\end{figure}

The introduced sensor model is validated and visualised in Figure~\ref{fig:sim_model}. Two projection rays, corresponding to the spherical and cylindrical regions, are depicted. Each ray intersects three relevant points: the frame of reference origin, the point in the 3D sensor surface, and the corresponding projected point in the image plane.

\subsection{Fabrication process}
\label{subsec:fabrication_process}

\begin{figure}
  \centering
  \includegraphics[width=\linewidth]{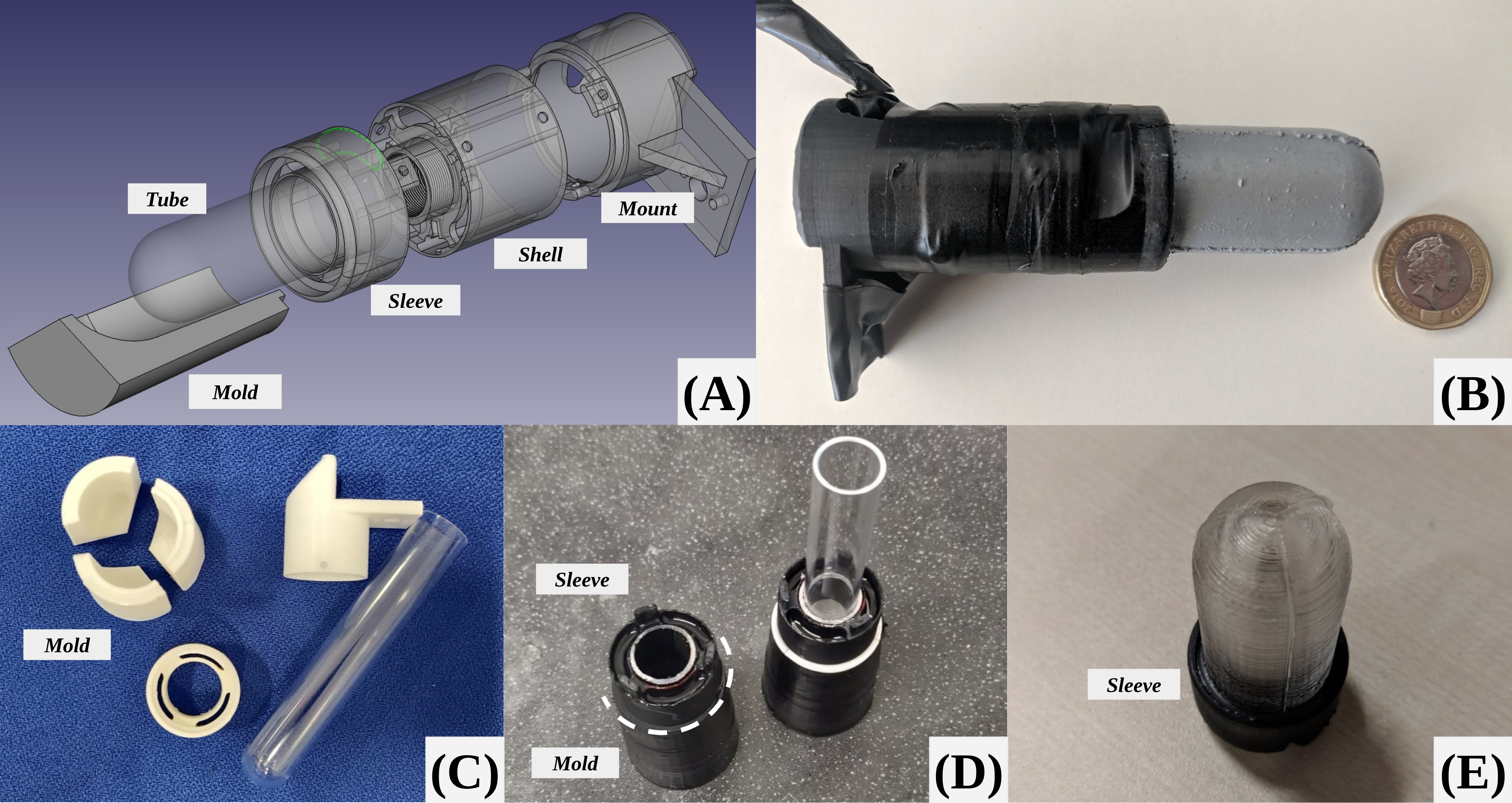} \\
  \caption{\textbf{(A)} Exploded view of the GelTip tactile sensor design.
  \textbf{(B)} A GelTip sensor, next to a British one pound coin, for relative size comparison. The sensor has a length of approximately \SI{10}{\cm}; its shell has a diameter of \SI{2.8}{\cm}; and the tactile membrane has a length of \SI{4}{\cm} and a diameter of \SI{2}{\cm}. 
 \textbf{(C)} The three-part \textit{mold} next to the remaining parts used in the GelTip construction. \textbf{(D)} The plastic tube is inserted into the sleeve and then mounted onto the mold, afterwards the tube is measured and trimmed and then the elastomer is poured. \textbf{(E)} The tactile membrane after being de-molded and before being painted. 
 }
 \label{fig:sensor_fabrication_proc}
\end{figure}

As for any other optical tactile sensors, the fabrication of the GelTip sensor is the fabrication of the sensing membrane. It requires the fabrication of three parts: the bottom rigid layer, the deformable elastomer, and the coat of paint. For constructing the rigid layer, a flat sheet of transparent acrylic can be used, in the case of flat sensors \cite{GelSight2017GeometrySlip}. However, for finger-shaped sensors, a curved rigid surface is necessary. In the case of the \textit{GelTip}, a simple off-the-shelf transparent test tube is used. These commercially available tubes are made of plastic/acrylic or glass and one disadvantage of using such test tubes, particularly the plastic ones, is that they contain small imperfections that result from the manufacturing process. An alternative approach is to print the rigid tube using a stereolithography 3D-printer and clear resin, however, proper polishing is necessary to ensure its optical transparency \cite{br2020soft}. To fabricate the elastomer in the desired shape a mold can also be created, for instance 3D printed. Fused Filament Fabrication (FFF) and Stereolithography (SLA) printers, yield different textured surfaces and consecutive differently textured elastomers. Example 3D-printed parts are shown in Figure~\ref{fig:sensor_fabrication_proc}-(C). 

The soft elastomer is then created by mixing a two-components silicone, such as XP-565. Mixing these two parts in different ratios yields elastomers with different a elastic properties. Additional additives can also be considered, such as, Slacker for increasing the silicone tackiness. A commonly used mixture used for the \textit{GelTip} would be 1 gram of XP-565 part-A, 22 grams of XP-565 part-B and 22 grams of the Slacker.

For painting the transparent elastomer, off-the-shelf spray paints tend to form a rigid coat and cracks will develop in the coat when the elastomer deforms or stretches. To avoid these issues, custom paint can be fabricated and applied using a paint gun or an airbrush \cite{GelSight2017}. Pigment powder is mixed with a small portion of \textit{part-A} and \textit{part-B} of XP-565, with the same ratio as used in the elastomer. The paint pigment commonly consists on aluminium powder (\SI{1}{\micro\metre}), however, other options can also be considered as well. After mixing them properly, the mixture is dissolved using a silicone solvent until a  watery liquid is achieved, that then can be sprayed onto the elastomer.

Finally the three sets of LEDs can be soldere: either of different colors, red, green and blue, or all white. Since different LEDs emit different light intensities, each cluster is preceded by a independent resistor. The power source can be either extracted from the camera or from an external source, e.g., adding a secondary USB cable. At the core of any optical sensor, a camera is installed. In the case of the \textit{GelTip}, a wide angle lens is also considered, enabling the recording of the internal surface of the entire finger.

\section{Evaluation}
\label{sec:experimentresults}
In this section, we look into two sets of experiments carried out using a \textit{GelTip} sensor. \review{The first set of experiments demonstrates how an image-based tactile sensor can be used to localise contacts and; the second set of experiments illustrates the advantages of leveraging all-around touch sensing for dexterous manipulation. Video recordings of these experiments and the CAD models for 3D printing the \textit{GelTip} sensor can be found at https://danfergo.github.io/geltip/}.


\subsection{Contact localisation}
\label{subsec:contact_detection}

\begin{figure}
\centering
\includegraphics[width=\linewidth]{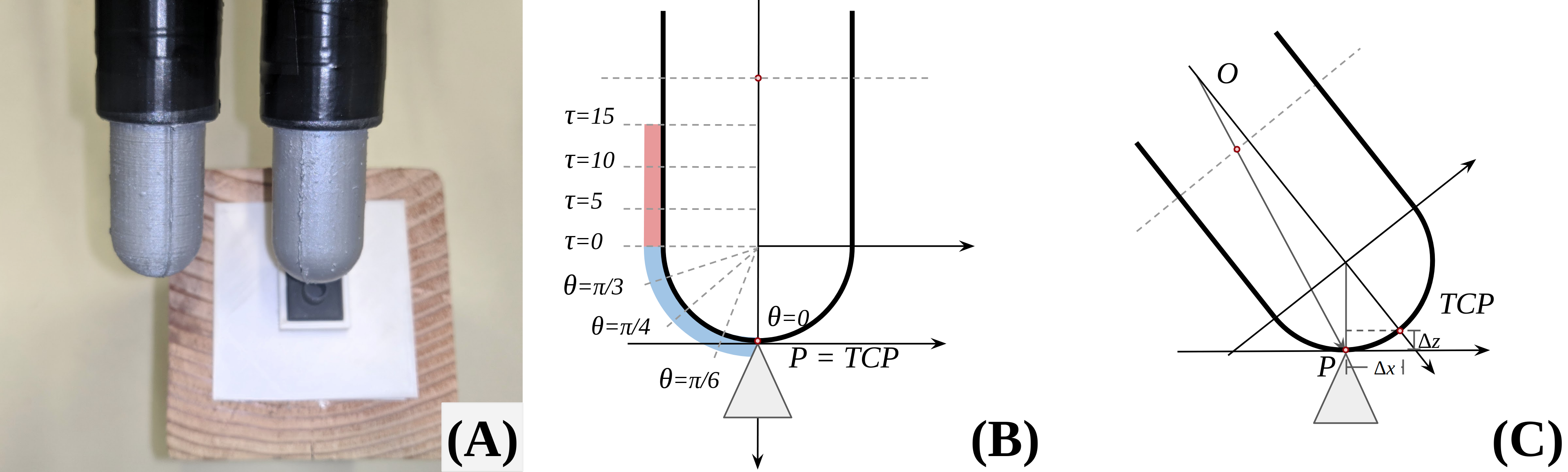} 
\caption{
\textbf{(A)} Two GelTip sensors are installed on a robotic actuator and a 3D printed mount that holds a small 3D printed shape (a cylinder here) placed on top of a wooden block. The actuator moves in small increments and collects tactile images annotated with the known contact positions.
\textbf{(B, C)} Illustration of the motion of the sensor during the data collection. The sensor starts pointing downwards, as shown in \textbf{(B)}. To obtain contacts on the sensor surface, while moving, the sensor is also translated by ($\Delta x$,~$\Delta z$), as shown in \textbf{(C)}. A total of eight contacts are collected per object: four rotations ($\theta$) on the sensor tip and four translations ($\tau$) on the sensor side, as highlighted in \textbf{(B)}.}
\label{fig:experiment_contact_detection}
\end{figure}

For the contact localisation experiment, a set of seven small objects is 3D printed, each with object with a maximum size of $1\times1\times2$ cm$^3$. The objects are shown in Table~\ref{table:contact_errors_per_object} . A 3D printed mount is also built, and placed on top of a raised surface, ensuring that all the objects are kept in the same position throughout the experiment. \textit{GelTip} sensors are installed on a robotic actuator, i.e., the 6-DoF Universal Robots UR5 arm with a Robotiq 2F-85 gripper. The actuator rotates and translates to tap objects at multiple known positions of one of its fingers surface, as illustrated in Figure~\ref{fig:experiment_contact_detection}.
The actuator starts with its fingers pointing downwards, i.e., orientation 0, being is visually aligned with the cone object. Contacts are then registered, firstly on the sensor tip, by rotating the sensor, and then on the side, by translating the sensor. In Figure~\ref{fig:experiment_contact_detection}~(B) markings show the location of such contacts and in Figure~\ref{fig:experiment_contact_detection}~(C) the necessary ($\Delta x$, $\Delta z$) translation to obtain contacts on the finger skin is also shown.
To use the projection model described in Section~\ref{sec:sensormodel}, five parameters are necessary to be known: $r$, $d$, $c_x$, $c_y$ and $\alpha$. The first two are extracted from the dimensions of the sensor design, however, the latter three are the intrinsic parameters of the camera, which need to be calibrated. To this end, we obtain such parameters from a known  pair of corresponding $(x',y')$ and $(x,y,z)$ points. We set the actuator to tap the object in the \SI{15}{\mm} translation position. The center of the sensor tip ($c_x, c_y$) and the contacted point are manually annotated in the image space. The $\alpha$ parameter can then be derived by fitting the known information into Equation \ref{eq:proj_ray}. 
After detecting the contact in the image space and projecting it into $(x,y,z)$ coordinates, the Euclidean distance between the predicted and the true contact positions are computed. For each of the seven objects, a total of eight contacts are recorded, i.e., four rotations ($\theta$): ~$0$, $\pi/6$, $\pi/4$, $\pi/3$; and four translations ($\tau$): \SI{0}{\mm}, \SI{5}{\mm}, \SI{10}{\mm}, \SI{15}{\mm}.  


The resulting localisation errors between the observed and true localisation, expressed in millimeters, are summarised in Table~\ref{table:contact_errors_per_position}. Overall, the variance of the localisation errors is large: in some contacts the obtained errors are lower than \SI{1}{\mm}, while in others are over \SI{1}{\cm}. On the other hand, the localisation error, for each object or position, is correlated with its variance. The largest localisation errors happen on objects with large or rounded tops i.e., sphere, edge and slab; contrariwise, the lowest errors are observed for objects with sharp tops, i.e., cone, tube and cylinder. In terms of the localisation errors at different positions, contacts happening near the sensor tip, i.e., the rotations, present lower errors than contacts happening on the sensor side, i.e., translations. In particular, contacts happening at $pi/4$ and $pi/6$ have the lowest errors. 

\begin{figure}
  \centering
  \includegraphics[width=\linewidth]{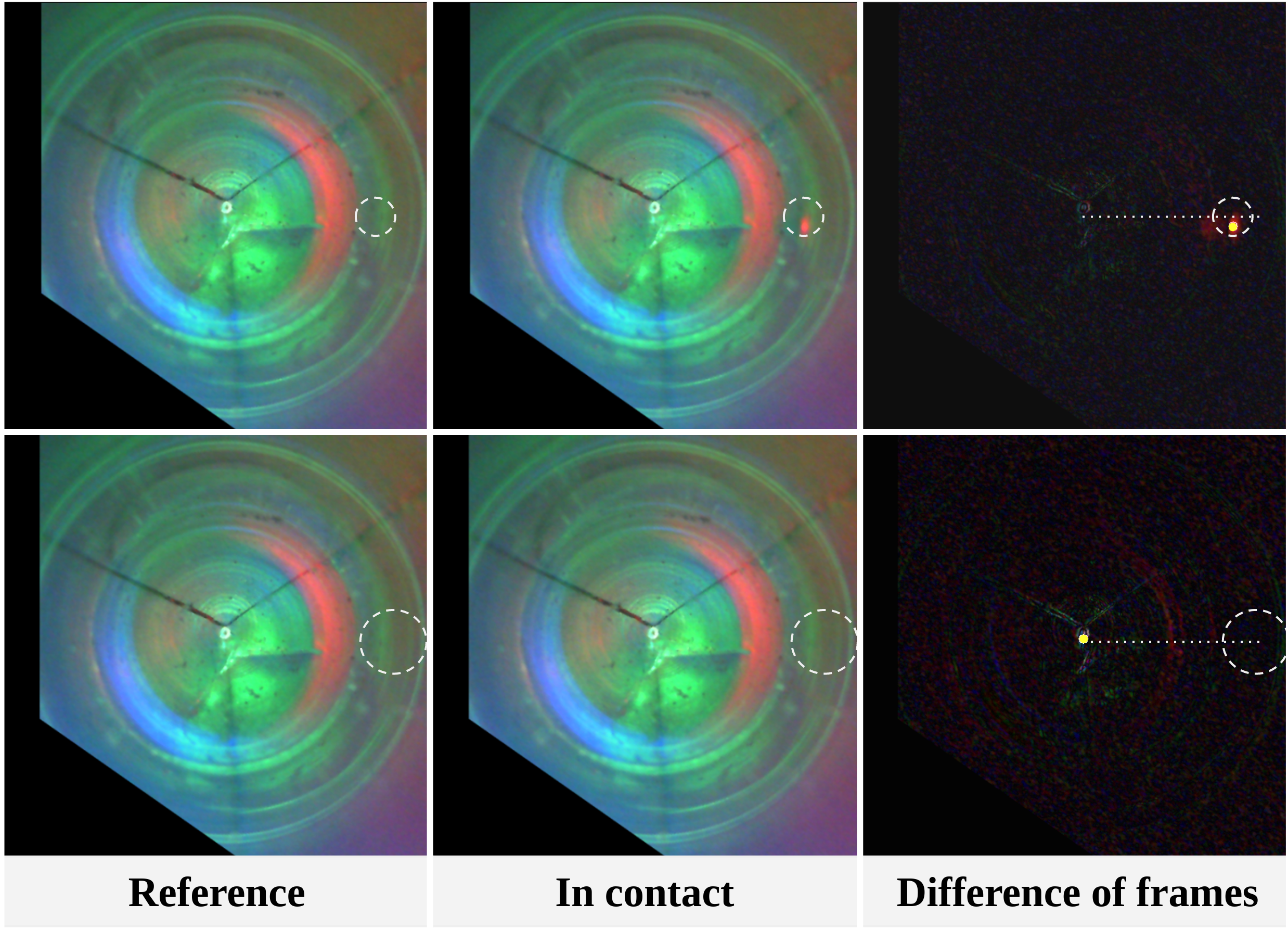} \\
\caption{\review{Reference and in contact frames for two evaluated contacts. The expected contact region is highlighted with a dash circumference, the predicted contact position with a yellow circle and the axis where the contacts occur with dotted lines. The top row shows the contact for the $<Cone, \theta=0>$ that results in the smallest error, and the bottom row shows the  $<Slab, \tau=\SI{15}{\mm}>$ that results in the largest error.}}
  \label{fig:contact_localisation_samples}
\end{figure}

\review{From these experiments, we find three main challenges using a finger-shaped sensor, such as the \textit{GelTip}, for contact localisation: 1) weak imprints, created by light contacts, may not be captured by the localisation algorithm; 2) forces perpendicular to the main axis of the sensor may flex the sensor tip, resulting in localisation errors; and 3) imperfections in the sensor modeling and calibration further contribute to these localisation inaccuracies. Examples of captured tactile images and corresponding predictions for the smallest (i.e., $<Cone, \theta=0>$) and largest localisation errors (i.e., $<Slab, \tau=\SI{15}{\mm}>$) are shown in Figure~\ref{fig:contact_localisation_samples}. In the first case, due to the bright imprint provided by the sharp cone top, the algorithm successfully locates the contact. In the second case, due to the imperceptible contact imprint the algorithm incorrectly predicts the contact in the sensor tip.}

\begin{table}
\setlength{\tabcolsep}{3.9pt}
\def\arraystretch{2.2}
\centering
\caption{Contact errors per position, expressed in millimeters}
\label{table:contact_errors_per_position}
\begin{tabular}{ccccccccc}
\Xhline{3\arrayrulewidth}
\multicolumn{4}{c}{\textbf{ROTATIONS}} & \multicolumn{4}{c}{\textbf{TRANSLATIONS}} \\[-1.5ex]
$0$ & $\pi/6$ & $\pi/4$ & $\pi/3$ & 0 & 5 & 10 & 15 \\ 
\Xhline{1.5\arrayrulewidth} 
\makecell{$4.71$\\$\pm0.75$} & 
\makecell{$2.01$\\$\pm0.90$} & 
\makecell{$1.04$\\$\pm0.46$} & 
\makecell{$6.96$\\$\pm4.82$} & 
\makecell{$7.87$\\$\pm5.08$} & 
\makecell{$8.03$\\$\pm1.92$} & 
\makecell{$7.55$\\$\pm5.00$} & 
\makecell{$4.86$\\$\pm8.41$} \\ 
\Xhline{3\arrayrulewidth} 
\end{tabular}
\end{table}

\begin{table}
\setlength{\tabcolsep}{5.2pt}
\def\arraystretch{2.2}
\centering
\caption{Contact errors per object, expressed in millimeters}
\label{table:contact_errors_per_object}
\begin{tabular}{cccccccccc}
\Xhline{3\arrayrulewidth}
\pbox{0.093\linewidth}{\includegraphics[width=0.95\linewidth]{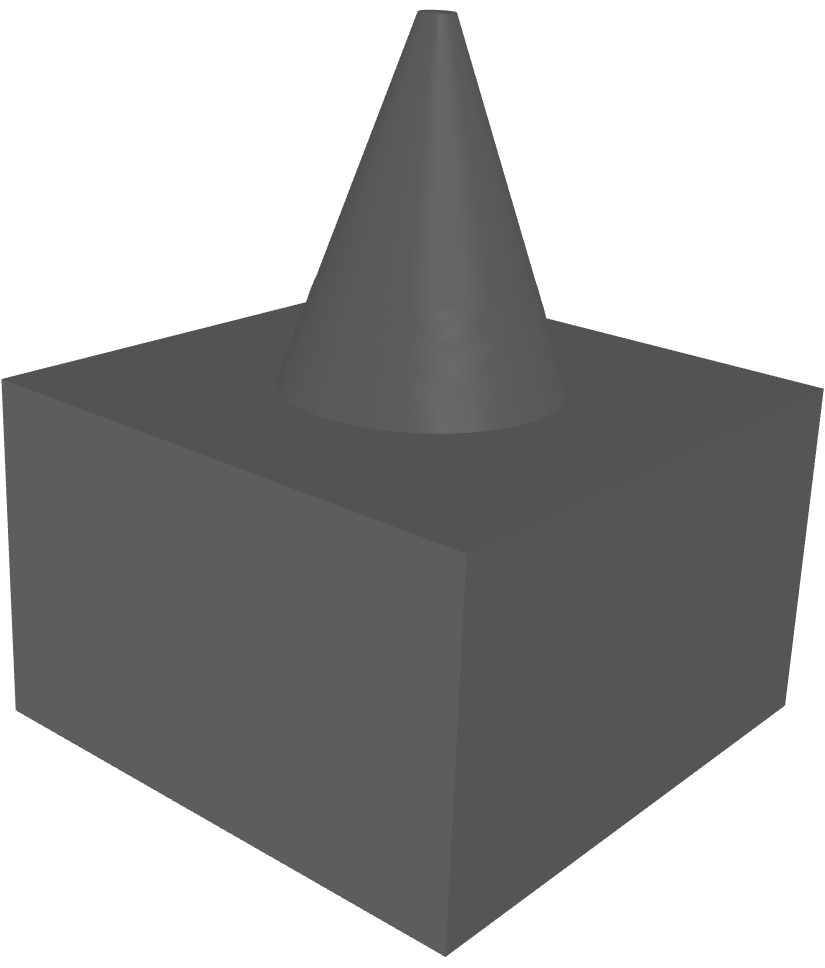} \\ \centering\textbf{Cone} }&%
\pbox{0.096\linewidth}{\includegraphics[width=\linewidth]{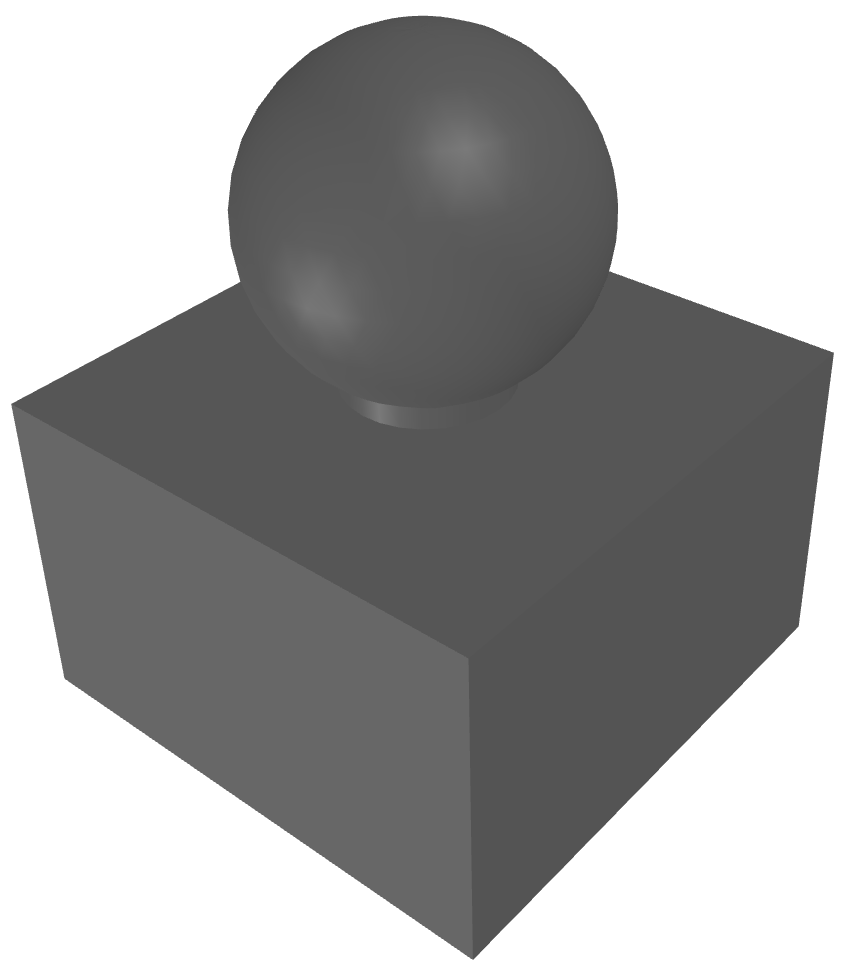} \\ \centering\textbf{Sphere} }&%
\pbox{0.12\linewidth}{\includegraphics[width=0.8\linewidth]{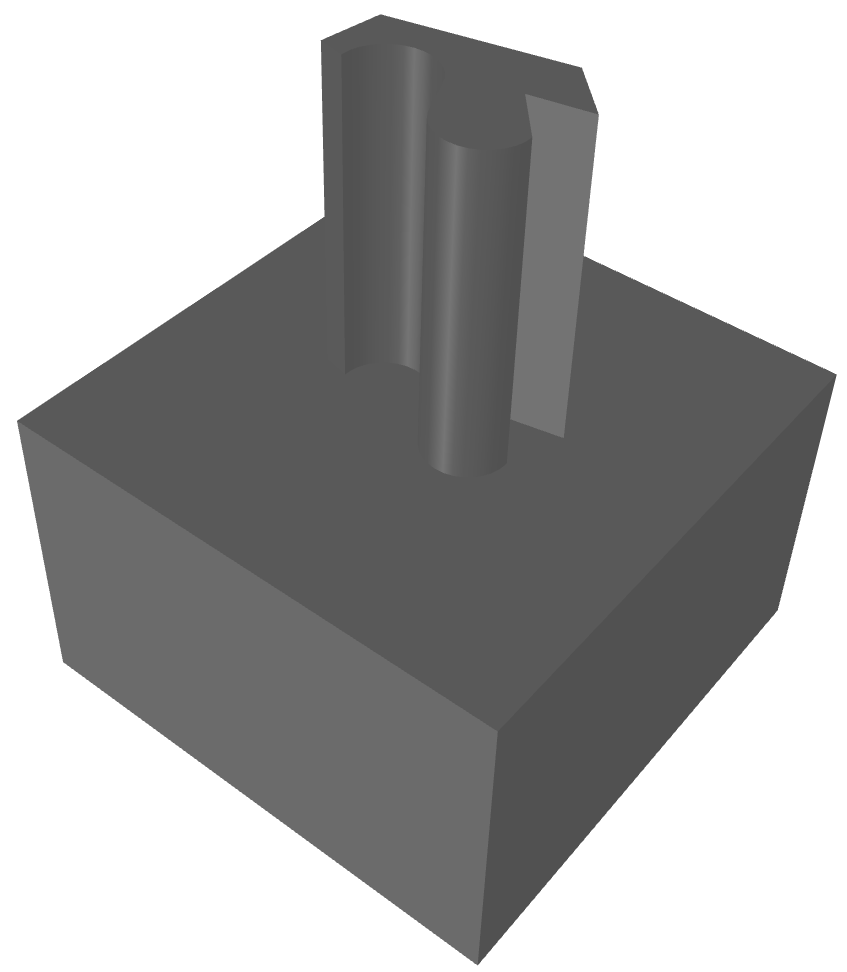} \\ \centering\textbf{Irregular} }&%
\pbox{0.094\linewidth}{\includegraphics[width=\linewidth]{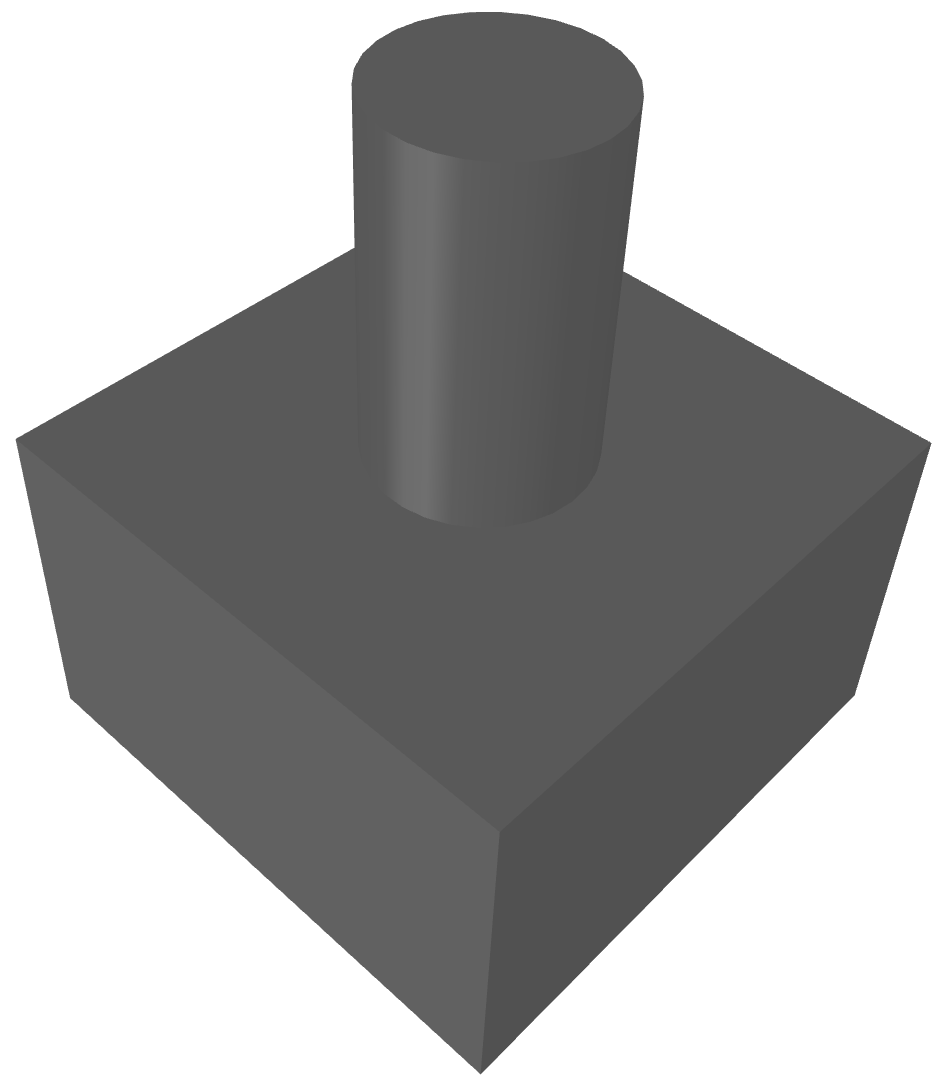} \\ \centering\textbf{Cylinder} }&%
\pbox{0.093\linewidth}{\includegraphics[width=\linewidth]{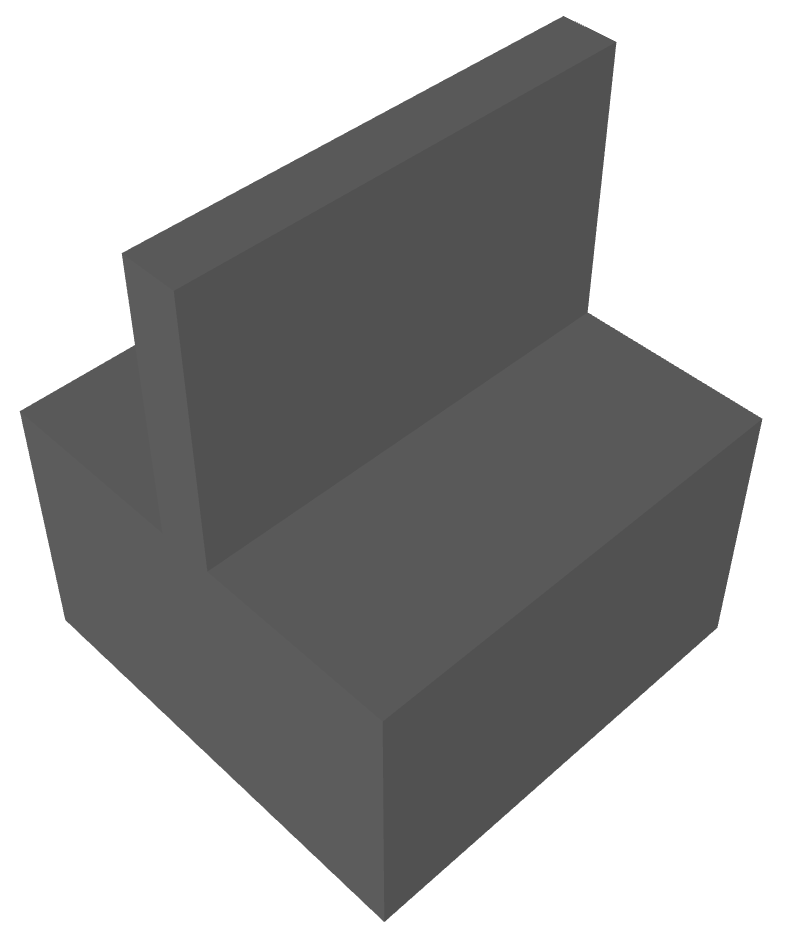} \\ \centering\textbf{Edge} }&%
\pbox{0.093\linewidth}{\includegraphics[width=\linewidth]{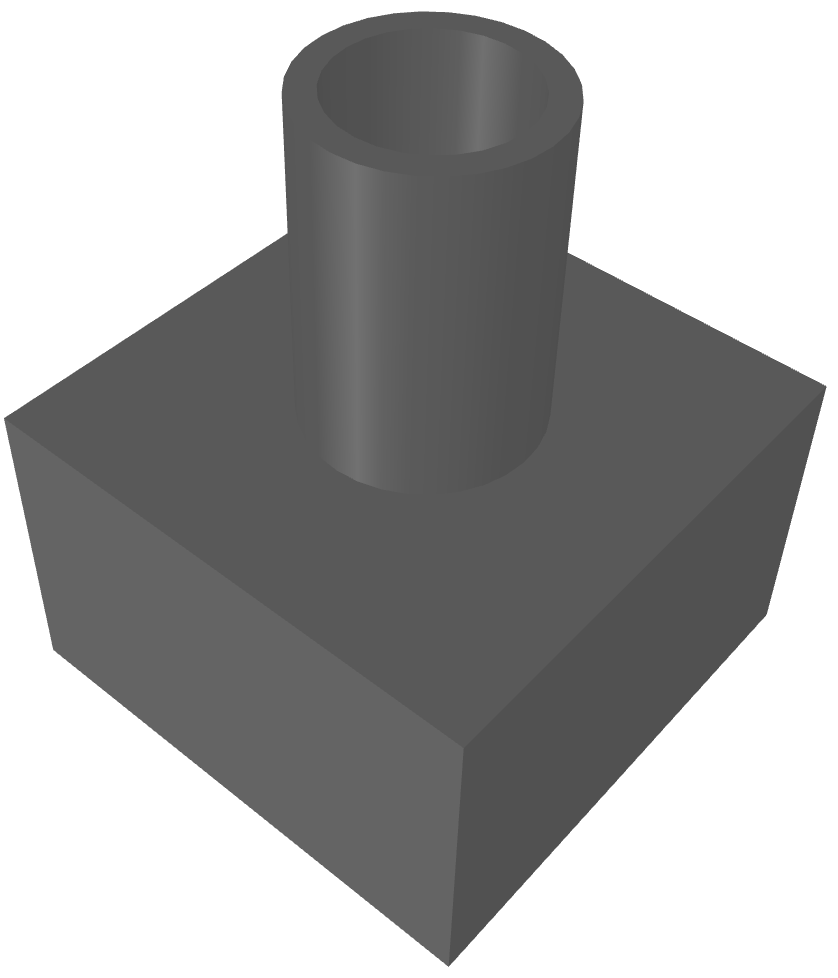} \\ \centering\textbf{Tube} }&%
\pbox{0.093\linewidth}{\includegraphics[width=0.9\linewidth]{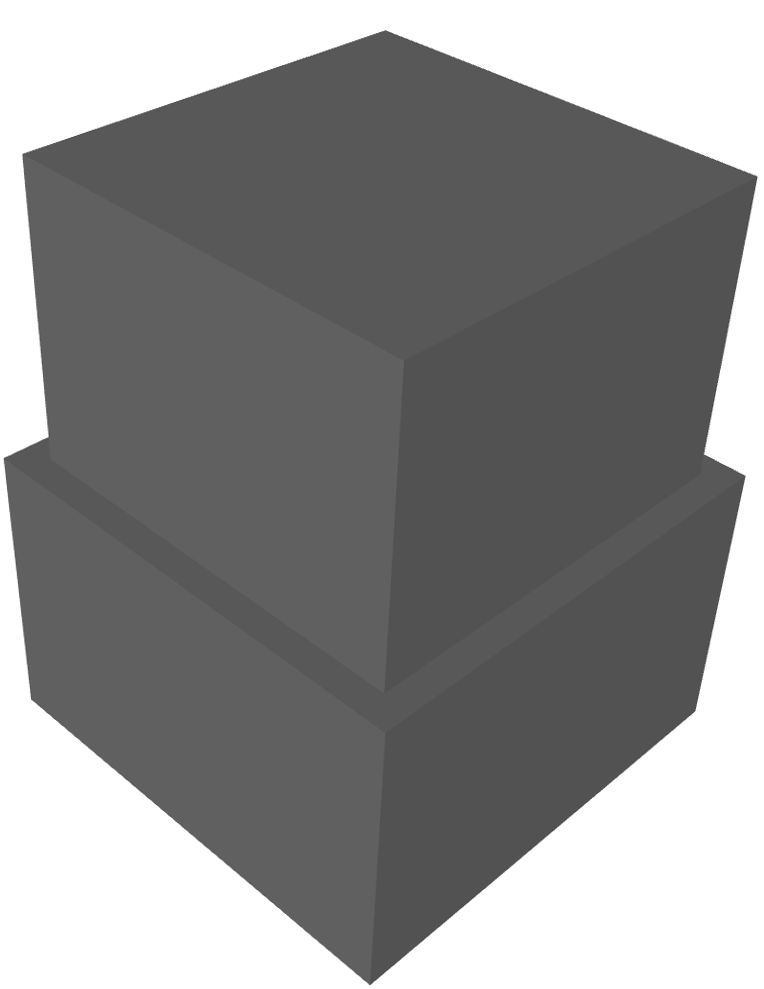} \\ \centering\textbf{Slab} } \\ 
\Xhline{1.5\arrayrulewidth} 
\makecell{$3.63$\\$\pm3.26$} & 
\makecell{$6.79$\\$\pm5.38$} & 
\makecell{$5.61$\\$\pm4.08$} & 
\makecell{$4.57$\\$\pm4.30$} & 
\makecell{$7.47$\\$\pm6.29$} & 
\makecell{$3.33$\\$\pm1.90$} & 
\makecell{$6.27$\\$\pm8.17$} \\
\Xhline{3\arrayrulewidth} 
\end{tabular}
\end{table}

\subsection{Touch-guided grasping in a Blocks World environment}
\label{sec:exp_blocks_world}
In the task of grasping objects, the initial motion of the gripper is often planned using remote sensing, e.g., camera vision or Lidar. However, remote sensing suffers from occlusions and inaccurate predictions about geometry and distances. In such cases, the final grasp and re-grasp control policies have to rely on inaccurate information of where and when contacts occur. In contrast, touch sensing offers accurate feedback about such contacts.

\begin{figure}
\centering
\includegraphics[width=0.5\linewidth]{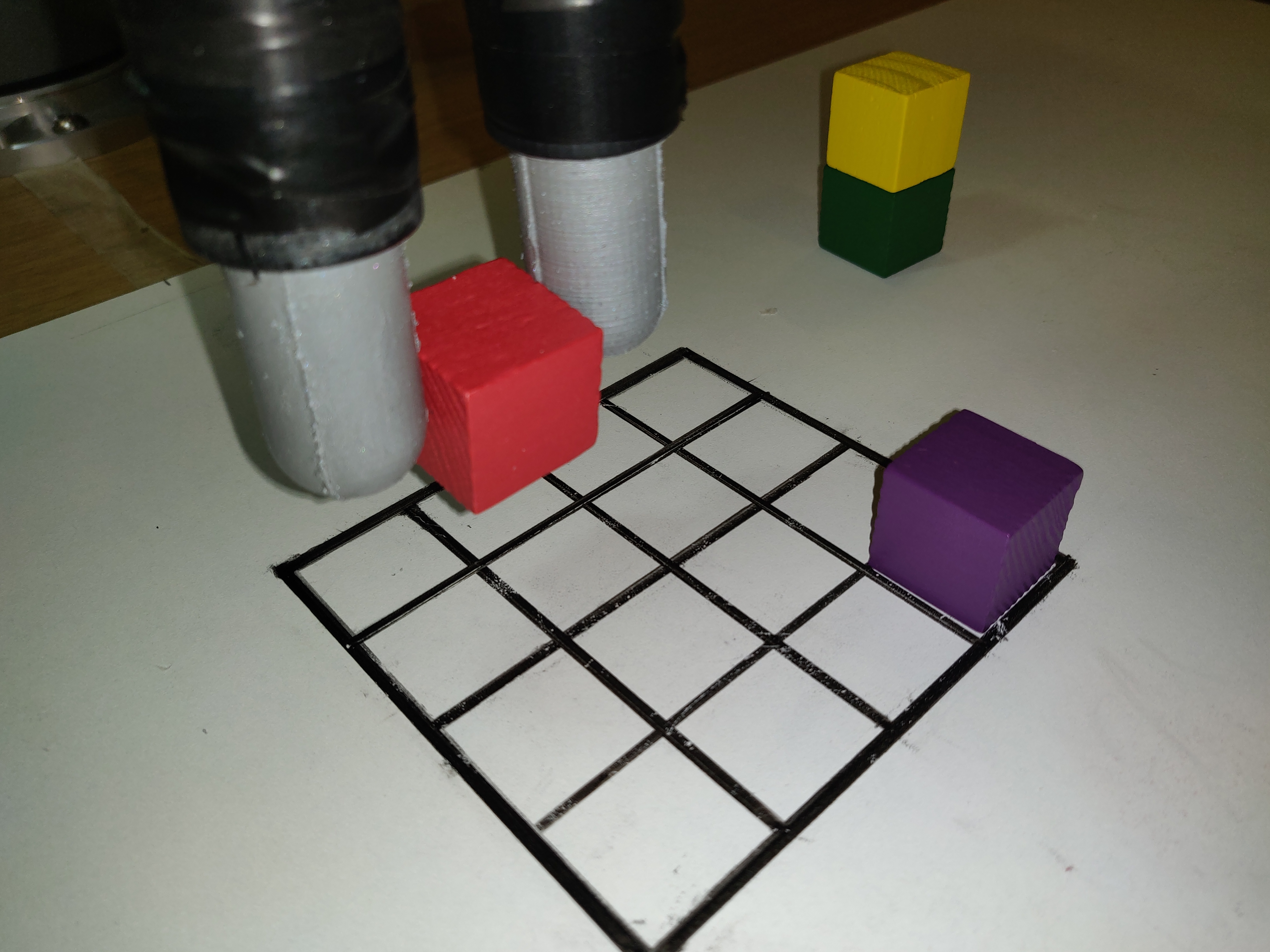} \\
\caption{The Blocks World experimental setup. In each experiment, the robot actuator moves row by row, attempting to grasp each block. The experiment shows that, even with the initial uncertainty, the robot grasps all the blocks successfully, using the all-around touch feedback.}
\label{fig:exp_blocks_world}
\end{figure}

We can clearly verify the importance of considering of touch sensing, by studying two different policies in a simple grasp experiment, i.e., \textbf{\textit{Random grasp}}  (Rg) and \textbf{\textit{Random grasp + Touch informed re-grasp}} (RgTr), in a simple grasp experiment. In this experiment, the robot is presented with a 4x4 board with one wooden block placed in each row in unknown columns (to the robot). The robot attempts to grasp each block, and if it fails to grasp one block after 5 attempts, it is considered a failure, and it skips to the next one, as shown in Figure~\ref{fig:exp_blocks_world}. 
Here, the random grasp in Rg and RgTr mimics the inaccuracies from remote sensing based grasping, by sampling the block position randomly. The touch informed re-grasp in RgTr, mimics an adaptation carried out by touch sensing, by sensing possible collisions, and moving the gripper towards the column in which the contact is detected. A \textbf{\textit{Control policy}} (C) can also be implemented for reference. In this case, the agent always knows the position of each block and consequently always moves directly towards it. 
The results of this experiment are summarised in Table~\ref{table:grasping_results}, after executing these policies 5 times. As it can be seen, in all the measured metrics, RgTr is a more successful policy than Rg. For instance, Rg fails to grasp 20\% of the blocks, i.e., on average one block is left on the board at the end of each run. In contrast, with the RgTr policy all the blocks are grasped, resulting in a failure rate of 0\%. Similarly, both the average number of attempts and the average number of collisions per block with the RgTr policy is also lower than the Rg policy, i.e., $1.85$ and $0.55$ \textit{versus} $3.30$ and $1.45$. This difference in performance is justified by the fact that in the case of RgTr, once a collision occurs the regrasp policy ensures that the grasp attempt is successful. If the grasp position is sampled randomly, it will be a success chance of $1/4$ for each grasping attempt. In contrast, with the touch feedback enabled, this chance jumps to $2.5/4$ on average. As a consequence, the RgTr policy finds a successful grasp more quickly and thus grasps more blocks within the maximum 5 attempts limit. This experiment shows that sensing contacts outside the grasp closure offers an important feature to improve the success chance of a given grasp attempt.

\begin{table}
\centering 
\caption{The table summarises the percentage of failing to grasp blocks (failure rate), and the average number of attempts and collisions per block in all the grasping attempts ($4 \times 5$). It can be noted that the \textit{Random Grasp + Touch Informed Regrasp} policy outperforms \textit{Random Grasp} in all three metrics, i.e., obtains lower failure rate, less attempts and less collisions.}
\def\arraystretch{1.5}
\begin{tabular}{|L{3.5cm}|p{2cm}|p{2cm}|p{2cm}|}
\hline
\backslashbox{\\Policies}{}& Failure rate & \# of attempts per block & \# of collisions per block \\ \hline\hline
\textbf{Control}         & 0\%   & 1          &      0 \\ \hline
\textbf{Random Grasp}               & 20\%  & 3.30       &   1.45 \\ \hline
\textbf{Random Grasp + Touch Informed Regrasp}             & 0\%   & 1.85       &   0.55 \\ \hline
\end{tabular}
\label{table:grasping_results}
\end{table}

\begin{figure}
  \centering
  \includegraphics[width=\linewidth]{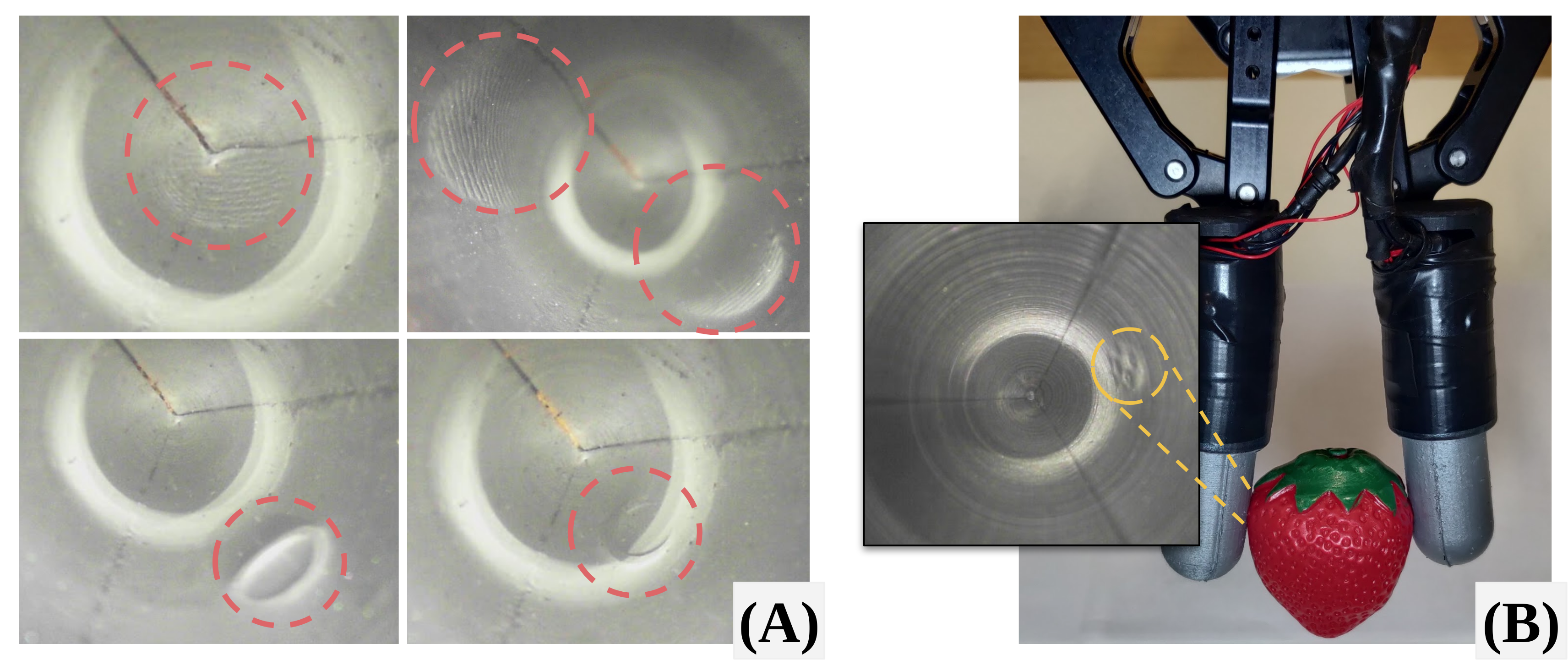} \\
\caption{\textbf{(A)} Tactile images captured using our proposed GelTip sensor. From left to right, top to bottom: a fingerprint pressed against the tip of the sensor, two  fingerprints on the sides, an open-cylinder shape being pressed against an side of the sensor and the same object being pressed against the corner of the tip, all highlighted in red circles. \textbf{(B)} A plastic strawberry being grasped by a parallel gripper equipped with two GelTip sensors, with the corresponding imprint highlighted in the obtained tactile image (in gray-scale).}
  \label{fig:sensor_demo}
\end{figure}

\section{Conclusions and Discussions}
\label{sec:conclusion}
In this chapter, we have reviewed the tactile sensors for robot grasping and manipulation, highlighted our proposed GelTip sensor, which can detect contacts around the robot finger. As illustrated in Figure~\ref{fig:sensor_demo}, it can capture fine ridges of human fingerprints and the fine texture of a plastic strawberry. The grasping experiments in the \textit{Blocks World} environment show the potential of the all-around finger sensing in facilitating dynamic manipulation tasks. In our future research, we will introduce imprinted markers to the GelTip sensor to track the force fields. The use of the GelTip sensor in the manipulation tasks, such as grasping in cluttered environments, will also be of our interest.

Compared to the GelSight sensors~\cite{GelSightSmallParts, GelSight2017}, due to the sensor design of a finger shape, the light distribution throughout the sensor internal surface is no longer homogeneous. Specifically, a brightly illuminated ring can be observed near the discontinuity region (see Figure~\ref{fig:sensor_innerworkings_and_model}~(B)). Shadows can also be observed in the bottom-left sample of Figure~\ref{fig:sensor_demo}~(A) when contacts of large pressure are applied, due to the placement of the camera and light sources. It may pose a challenge to geometry reconstruction using the Poisson reconstruction method \cite{GelSightSmallParts, GelSight2017, br2020soft} that builds a fixed mapping of pixel intensities to surface orientations and requires carefully placed RGB LEDs. In future research, Convolutional Neural Networks could be used for geometry reconstruction of the GelTip sensor. 


\section*{ACKNOWLEDGMENT}
This work was supported by the EPSRC project ``ViTac: Visual-Tactile Synergy for Handling Flexible Materials'' (EP/T033517/1).
\Backmatter

\chapter*{References}
\markboth{References}{References}
\bibliographystyle{elsarticle-num}
\bibliography{references}

\end{document}